\documentclass[balance,hyphenate]{asmejour}

\usepackage{tabularx}
\usepackage{amsmath} 
\usepackage{algorithmic}
\usepackage{algorithm}
\usepackage{graphicx}
\usepackage{textcomp}
\usepackage{xcolor}

\hypersetup{
	pdfauthor={John Shim, Christoph P. O. Nuesslein, Sixu Zhou, Hanjun Kim, Kinsey Herrin, Aaron J. Young},
	pdftitle={End-to-End Control of a Powered Knee-Ankle Prosthesis Towards Unified, Tuning-Free Assistance},
	pdfkeywords={Powered prosthesis, deep learning, end-to-end control},
	pdfsubject={End-to-end deep learning control of a powered knee-ankle prosthesis},
}

\JourName{Dynamic Systems, \\ Measurement, and Control}

\begin{document}

\SetAuthorBlock{John Shim\CorrespondingAuthor}{Woodruff School of Mechanical Engineering, \\
Institute for Robotics and Intelligent Machines, \\
Georgia Institute of Technology, \\
Atlanta, GA 30332-0405 USA \\
email: jshim77@gatech.edu}

\SetAuthorBlock{Christoph P. O. Nuesslein}{Woodruff School of Mechanical Engineering, \\
Institute for Robotics and Intelligent Machines, \\
Georgia Institute of Technology, \\
Atlanta, GA 30332-0405 USA}

\SetAuthorBlock{Sixu Zhou}{Woodruff School of Mechanical Engineering, \\
Institute for Robotics and Intelligent Machines, \\
Georgia Institute of Technology, \\
Atlanta, GA 30332-0405 USA}

\SetAuthorBlock{Hanjun Kim}{Woodruff School of Mechanical Engineering, \\
Georgia Institute of Technology, \\
Atlanta, GA 30332-0405 USA}

\SetAuthorBlock{Kinsey Herrin}{Woodruff School of Mechanical Engineering, \\
Georgia Institute of Technology, \\
Atlanta, GA 30332-0405 USA}

\SetAuthorBlock{Aaron J. Young}{Woodruff School of Mechanical Engineering, \\
Institute for Robotics and Intelligent Machines, \\
Georgia Institute of Technology, \\
Atlanta, GA 30332-0405 USA}

\title{End-to-End Control of a Powered Knee-Ankle Prosthesis Towards Unified, Tuning-Free Assistance}


\begin{abstract}
Powered prostheses conventionally rely on impedance controllers requiring extensive manual tuning and explicit mode classification. In this work, we present the real-time deployment of a prosthesis controller mapping onboard sensor signals directly to continuous actuator commands, replacing both the intent classifier and the impedance parameterization with a learned policy. Temporal Convolutional Networks were trained on a multi-terrain dataset from 18 individuals with transfemoral amputation. Four participants (three able-bodied via bypass adapter, one with transfemoral amputation) ambulated across five locomotion modes with the learned policy in real-time. During level walking, the controller scaled peak ankle torque with speed in the same direction as its training data (0.52 deployed, 0.96~Nm/kg per m/s training); excluding one participant with atypical loading sharpened this to 0.85. Feature
attribution identified the sagittal load cell moment as the dominant driver of ankle torque, accounting for 72\% of the torque increase above resting baseline, indicating the controller converged on a biomechanically informative speed cue. Mode-distinct behavior was produced across ramps and stairs. These preliminary results support continuous end-to end control as a route to unified, mode-adaptive assistance with reduced tuning burden.
\end{abstract}

\maketitle

\section{Introduction}

Individuals with transfemoral (TF) amputation (i.e., above-knee) routinely encounter limitations in functional mobility, exhibiting reduced walking speed, elevated metabolic cost, and degraded gait symmetry~\cite{vanvelzen2006, bona2020}. These deficits are exacerbated by passive prostheses, which cannot deliver net positive joint work, leading users to compensate by overloading their intact joints. This further elevates the risk of osteoarthritis and chronic joint pain~\cite{gailey2008}. Powered prostheses have the potential to close these functional gaps through actuated knee and ankle assistance, which has been shown to enable step-over-step stair ascent~\cite{camargo2023} and improved self-selected walking speed~\cite{herr2011}.

\begin{figure*}[t]
\centering
\includegraphics[width=\textwidth]{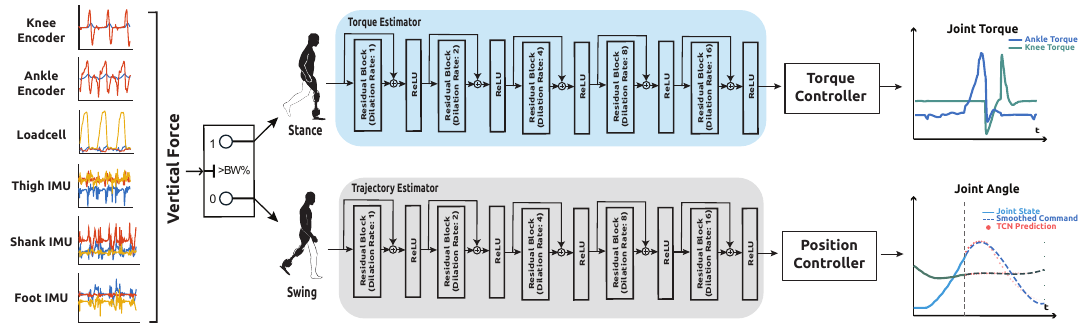}
\caption{Proposed end-to-end control framework. A deep neural network maps onboard sensor measurements—joint encoders (position, velocity), IMU (3-axis acceleration, angular velocity), and load cell (3-axis force, moment)—to continuous joint torque and trajectory estimates in real time. The control mode is governed by the axial load cell reading exceeding a fraction of body weight. The shaded region indicates trajectory prediction.\label{fig:framework}}
\end{figure*}

Realizing the benefits of powered assistance requires coordinated control. This process is conventionally implemented through a hierarchical framework~\cite{sup2008}: a high-level controller infers user intent from sensor signals (e.g., locomotion mode~\cite{bhakta2025} and gait speed~\cite{kim2025}), a mid-level controller computes desired joint torques, typically via impedance control paired with a finite state machine (FSM)~\cite{simon2014, bhakta2020}, and a low-level controller realizes actuator commands. This paradigm faces two key scalability limitations despite its efficacy. First, impedance controllers require extensive tuning; accommodating varied locomotion modes and terrain conditions (e.g., stair heights or ramp slopes) can take trained clinicians up to five hours per user~\cite{simon2014}, making deployment prohibitively costly. Second, because impedance parameters are defined over a discrete set of locomotion modes, intent misclassification may trigger abrupt transitions in assistance. Individuals with lower-limb amputation take 3,000--4,000 steps daily~\cite{stepcount2017}; even a 99\% accurate classifier yields 30--40 misclassifications per day, each of which can destabilize users and increase fall risk~\cite{zhang2015}.

To improve deployment scalability, recent work has addressed these limitations from two angles. To reduce tuning effort, modern optimizers have been utilized to automatically derive impedance parameters. Wen \textit{et~al.}~\cite{wen2020} and Best \textit{et~al.}~\cite{best2023} leveraged online reinforcement learning and convex optimization, respectively. Best's work also partially addressed the discrete control paradigm by unifying ramp and walking conditions, though stair remained separate. The second angle pushes this idea further, pursuing fully \textit{unified control} that replaces mode-specific behaviors with a single classification-free controller. Sullivan \textit{et~al.}~\cite{sullivan2025} realized this through a motion-primitive framework that handles transitions without a mode classifier. These advances ease tuning or unify control individually; however, each remains constrained to predefined locomotion modes or requires controller-specific expertise for personalization, leaving scalable deployment unresolved.

Deep learning-based control offers a promising path to address both challenges. By generating actions directly from sensor data, it collapses the high- and mid-level hierarchy and removes the need for mode classification and FSMs~\cite{molinaro2024}. In the prosthesis domain, Kim \textit{et~al.}~\cite{kim2022} demonstrated this paradigm by training a deep neural network to predict expert-tuned impedance parameters in real time, enabling ambulation across five locomotion modes without explicit intent classification.

\begin{table*}[t]
\centering
\caption{Comparison of Control Strategies for Powered Knee-Ankle Prostheses\label{tab:related}}
\renewcommand{\arraystretch}{1.5}
\begin{tabularx}{\textwidth}{@{}XXXXll@{}}
\toprule
Author & Control target & Mode handling & Parameter derivation & Real-time deployment & Evaluation participants \\
\midrule
Simon \textit{et al.} \cite{simon2014} & Impedance & FSM + classifier & Heuristic tuning & Yes & 6 TF \\
Best \textit{et al.} \cite{best2023} & Impedance (stance) + position (swing) & Classifier & Convex optimization & Yes & 2 TF \\
Sullivan \textit{et al.} \cite{sullivan2025} & Impedance (stance) + position (swing) & Classifier-free & Heuristic tuning & Yes & 3 TF \\
Kim \textit{et al.} \cite{kim2022} & Impedance & FSM + classifier-free & Machine learning & Yes & 3 TF \\
Nuesslein \& Young \cite{nuesslein2024} & Torque (stance) & Classifier-free & Machine learning & Offline & --- \\
\textbf{This work} & \textbf{Torque (stance) + position (swing)} & \textbf{Classifier-free} & \textbf{Machine learning} & \textbf{Yes} & \textbf{3 AB + 1 TF} \\
\bottomrule
\end{tabularx}
\end{table*}

However, predicting impedance parameters constrains what the network can learn. Training targets remain tied to impedance formulations, and the discrete nature of certain parameters---such as the knee equilibrium angle being non-zero only during stairs~\cite{simon2014}---implicitly encodes mode-dependent behaviors that reintroduce misclassification risks. Table~\ref{tab:related} summarizes how these approaches differ in what the controller commands, how locomotion modes are handled, and how control parameters are derived.

To bridge this gap, this study evaluated a fully end-to-end controller driven by continuous estimates of actuator signals. Prior work demonstrated a Temporal Convolutional Network (TCN) capable of estimating adaptive joint torque by inferring environmental context from onboard sensor signals \cite{nuesslein2024}. However, this approach was restricted to stance-phase and was not validated online. Building on this work, the present study extends the framework (Fig.~\ref{fig:framework}) to include swing-phase control by training a TCN to predict forward joint position trajectories. We evaluated the real-time deployment of this combined framework across a variety of ambulation tasks. Two hypotheses guided this work: our first hypothesis was that a TCN trained end-to-end on continuous
actuator signals preserves the assistance-scaling relationships present in
its training data. Our second hypothesis was that the resulting controller produces
distinguishable joint behaviors across locomotion modes despite receiving
no explicit mode label.

The contributions of this work are as follows:
\begin{enumerate}
    \item We deploy an end-to-end prosthesis controller in closed loop that commands
    continuous joint torque and position across the full gait cycle, with
    no impedance parameterization, intent classifier, or gait-phase
    estimator.

    \item We introduce a swing-phase controller that tracks learned forward
    joint position trajectories, extending end-to-end actuator-signal
    control beyond the stance-only formulation of prior work
    \cite{nuesslein2024}.

    \item We show through feature attribution and targeted perturbation
    that the learned policy converges on a biomechanically informative
    loading cue without architectural priors.
\end{enumerate}


\section{Methods}
\subsection{Prosthesis Dataset}
\label{sec:dataset}

\begin{figure}[t]
\centering
\includegraphics[width=\columnwidth]{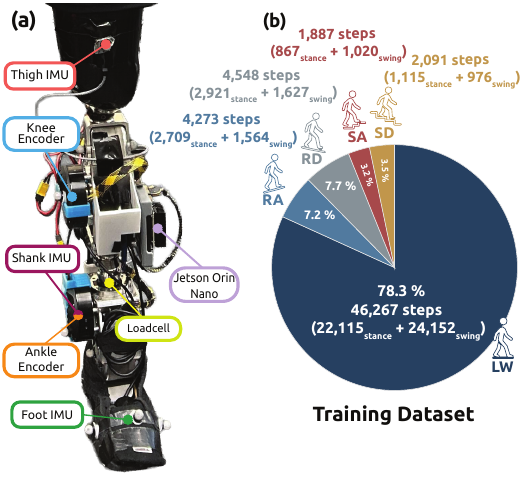}
\caption{(a) Onboard sensor configuration of the Open-Source Leg. (b) Training data distribution by locomotion mode, with stride counts shown for stance and swing dataset.\label{fig:hardware}}
\end{figure}

A multi-terrain locomotion dataset was compiled from prior experiments involving the OSL under real-time impedance control \cite{camargo2023, bhakta2025, bhakta2020, zhou2025, maldonado2025}. Data from 18 individuals with TF amputation ($82.0 \pm 16$~kg, $1.75 \pm 0.11$~m, 12R/6L) were collected, spanning level walking (LW, 0.4--0.9~m/s), ramp ascent (RA) and ramp descent (RD, $7.8$--$12.4^\circ$), and stair ascent (SA) and stair descent (SD, 10.2--17.8~cm), totaling 29{,}727 stance and 29{,}339 swing steps (Fig.~\ref{fig:hardware}(b)). Non-LW modes included transition steps to and from LW, with the intact side leading ascent modes and the prosthetic side leading descent modes. The dataset was aggregated longitudinally across multiple experimental protocols, resulting in heterogeneous within-mode control policies. For instance, some RA conditions used LW-equivalent impedance parameters~\cite{bhakta2020}, while others incorporated knee flexion at heel contact~\cite{zhou2025}. Separately, the axial load cell threshold governing the stance-to-swing transition, and the magnitude and timing of the late-stance knee flexion torque, were individually tuned per participant across experiments.

\subsection{Hardware}
The first- and second-generation Open-Source Leg (OSL)~\cite{azocar2020, best2024} were used to collect the locomotion dataset, with real-time evaluation on the second-generation device for its expanded ankle range of motion (30$^\circ$ to 60$^\circ$). Onboard sensors (Fig.~\ref{fig:hardware}(a)) include motor encoders (Dephy, Maynard, MA), a 6-axis load cell (M3564F, Sunrise Instruments, Canton, MI), and three 6-axis IMUs: one integrated within the ankle module (MPU9250, TDK InvenSense, San Jose, CA) and two mounted on the thigh socket and foot plate (3DMGX5-25, HBK World, N\ae rum, Denmark). All sensors were resampled to 100~Hz and synchronized through ROS2. Real-time inference ran on a Jetson Orin Nano (NVIDIA, Santa Clara, CA) mounted at the knee.

\subsection{Network Architecture and Training}

TCNs~\cite{bai2018} were implemented to cover the stance and swing modes. Dilated causal convolutions provide a long effective receptive field at low depth, enabling fast inference time suitable for real-time deployments. Both networks used kernel size $k=4$, depth $L=5$, and 64 filters per layer, yielding 1.87~s of effective history  (Fig.~\ref{fig:framework}). Hyperparameters were taken from prior work~\cite{nuesslein2024}.

The stance model input comprised 28 sensor channels; the swing model augmented this with a sagittal-plane thigh tilt angle from the thigh IMU. The stance model predicted knee and ankle torques with a 40~ms temporal lead, accounting for the group delay of the output low-pass filter (cut-off frequency of 15 Hz) and one 100 Hz control cycle. Torque labels were normalized by the user's body weight. The swing model predicted a 300~ms horizon of future joint positions, where the upper bound was set by the self-selected swing-phase duration of individuals with TF amputation walking with passive prostheses (520--570~ms~\cite{jaegers1995}) and constrained below by the accuracy degradation observed at longer horizons (Fig.~\ref{fig:swing_horizon}).

The training protocol used Adam optimizer with mean squared error (MSE) loss and early stopping. All sensor inputs were Z-score normalized so that channels of differing units and magnitudes contributed comparably to the gradient. To address mode imbalance, locomotion modes were resampled each epoch to a fixed distribution (60\% LW, 10\% per remaining mode). Performance was evaluated offline via leave-one-subject-out cross-validation.

\subsection{Torque and Kinematic Control}
The controller used two control states, selected by the contact condition. During stance, the limb is
kinematically constrained by ground contact and the prosthesis must
exchange force with the environment to influence leg behavior; torque
control was therefore selected. During swing, the limb moves
in free space, where the dominant requirement is foot clearance and
accurate positioning for the subsequent heel strike. Thus, direct position
control was considered desirable. The axial load cell reading determined the contact state, with thresholds at 25\% body weight (BW) for swing initiation and
15\% BW for stance entry.

This differs from prior state machine formulations in what the switch
encodes. Conventional controllers segment the gait cycle into multiple
phases and assign a distinct parameter set to each; here, within-phase behavior is produced continuously by the network, and the only discrete
decision is whether the limb is loaded. Unlike a locomotion mode
classifier, whose errors commit an entire stride to incorrect assistance, a
mis-triggered contact transition is a single-sample event on a directly
measured physical quantity and self-corrects at the next threshold
crossing.

During stance, the actuator's embedded closed-loop current controller executed commanded torques, with two supplementary damping terms added for user safety:
\begin{equation}
    \tau_{cmd} = \tau_{TCN} + \tau_{damp} + \tau_{wall}
\end{equation}
where $\tau_{TCN}$ is the torque estimated by the TCN, $\tau_{damp}$ is a resistive damping torque to prevent knee buckling, and $\tau_{wall}$ is a one-sided viscous damper activating near the positional boundary. The damping torque
\begin{equation}
    \tau_{damp} = \max\left(\tau_{max},\, \min(0, B_d \dot{\theta})\right)
\end{equation}
provides a resistive extension torque proportional to joint angular velocity $\dot{\theta}$ ($B_d = 5$~Nm/rad/s, capped at $\tau_{max} = 10\%$ BW) to prevent knee buckling. Additionally, the wall torque
\begin{equation}
    \tau_{wall} = \begin{cases} B_w \dot{\theta}, & \theta < \theta_w \text{ and } \dot{\theta} < 0 \\ 0, & \text{otherwise} \end{cases}
\end{equation}
acts as a one-sided viscous damper near the joint's positional boundary, activating only during knee extension to prevent terminal impact. $B_w$ and $\theta_\omega$ represent wall damping gain and boundary angle, respectively. While these terms were not necessary to walk with the controller, they were incorporated and modulated based on the preference and comfort of the user.

During swing-phase, the controller utilized closed-loop position feedback with a velocity feedforward term. The trajectory estimates are smoothed at each inference step by fitting a third-order polynomial to the TCN-predicted horizon $\hat{p}_{1:30}$ using a constrained least-squares optimization. The polynomial is anchored to the previous commands ($p_0$, $v_0$) to enforce $C^1$ continuity. From the fitted trajectory, the position and velocity are evaluated at each control loop $\Delta t$ = 10 ms to generate the swing command. This process is outlined in Algorithm~\ref{alg:smoothing}.

\begin{algorithm}[t]
\caption{Polynomial Trajectory Smoothing\label{alg:smoothing}}
\begin{algorithmic}[1]
\REQUIRE $p_0, v_0, \hat{p}, t, \Delta t$
\STATE $y \leftarrow \hat{p} - (p_0 + v_0 t)$
\STATE $A \leftarrow [t^2, t^3]$
\STATE $[c_2, c_3] \leftarrow \arg\min_c \|Ac - y\|^2$
\item[] \textit{// Evaluate polynomial at next timestep}
\STATE $p^* \leftarrow p_0 + v_0 \Delta t + c_2 \Delta t^2 + c_3 \Delta t^3$
\STATE $v^* \leftarrow v_0 + 2 c_2 \Delta t + 3 c_3 \Delta t^2$
\end{algorithmic}
\end{algorithm}

\subsection{Experimental Protocol}
Controller evaluation was conducted on the OSL with three able-bodied (AB) participants ($28 \pm 5$ years, $81.5 \pm 7.5$~kg, $1.75 \pm 0.01$~m) and one TF participant (21 years, 65.7~kg, 1.65~m, left-side amputation). AB participants mounted the prosthesis on the right side via a bypass adapter (iWalk3.0, iWALKFree, Long Beach, CA). AB01 and AB03 had ambulated on the device with the bypass adapter across multiple prior sessions, while AB02 was new to the device. TF01 had prior experience with the powered prosthesis. The study was approved by the Georgia Institute of Technology Institutional Review Board under Protocol H21117 and all participants provided written informed consent before participating.

Level and ramp walking were evaluated on an instrumented split-belt treadmill (BERTEC, Columbus, OH). To assess speed adaptation, participants walked at 0.4 to 0.8~m/s in 0.1~m/s increments for one minute per condition. Ramp performance was assessed at 0.4~m/s across four inclines ($7.8^\circ$, $9.2^\circ$, $10.8^\circ$, $12.4^\circ$) in both ascent and descent. 

The proposed controller was additionally evaluated against a fixed baseline FSM controller, previously parameterized during the training data
collection. Participants walked at a fixed treadmill speed of 0.5~m/s, matching their pace to a metronome set at a cadence of their own preference. This was to keep gait-cycle timing consistent across trials and controllers. To assess the consequences of closed-loop deployment, the sensor signals recorded during the FSM baseline trial were additionally fed into the TCN offline, and compared against the network's real-time closed-loop output.

Transition behavior was evaluated on an instrumented terrain park with a five-step staircase (12.7~cm rise per step). Participants completed a circuit of level walking, ascent to the top platform with the intact side leading, turn-in-place, descent with the prosthetic side leading, and return to level walking.

\subsection{Sensor Feature Attribution}
Integrated Gradients (IG)~\cite{sundararajan2017} were applied to attribute the controller's predictions back to its sensor inputs along a path-integrated gradient. For a target output $f(\mathbf{x})$ and input $\mathbf{x} \in \mathbb{R}^{C \times T}$ ($C=28$ channels, $T=187$ time steps), IG attributes each input feature $(c, t)$ relative to a baseline $\mathbf{x}'$ as
\begin{equation}
    \mathrm{IG}_{c,t}(\mathbf{x}) = (x_{c,t} - x'_{c,t}) \int_{0}^{1} \frac{\partial f\!\left(\mathbf{x}' + \alpha(\mathbf{x} - \mathbf{x}')\right)}{\partial x_{c,t}} \, d\alpha,
\end{equation}
approximated via a 100-step midpoint Riemann sum along the straight-line path from $\mathbf{x}'$ to $\mathbf{x}$. The baseline $\mathbf{x}'$ was set to the cross-subject mean sensor reading during quiet standing. Per-channel attributions were obtained by summing along the time dimension,
\begin{equation}
    \mathrm{IG}_c(\mathbf{x}) = \sum_{t=1}^{T} \mathrm{IG}_{c,t}(\mathbf{x}),
\end{equation}
and averaged across trials to summarize per-feature contributions to a given prediction target.

\subsection{Statistical Analysis}
Stride-level linear mixed-effects models (LMMs) with subject random effects were used to test trends across continuous predictors (walking speed, ramp grade). Denominator degrees of freedom ($df$) for fixed-effect tests were determined using the Satterthwaite approximation to account for the correlated stride-level observations.

To test whether the TCN reproduced the training-data scaling pattern during LW, peak ankle torque was modeled as $\text{peak torque} \sim \text{speed} + (\text{speed} \mid \text{subject})$, fit independently to the training and deployed data; fixed-effect estimates and $95\%$~CIs were compared side-by-side.

Two ramp mode contrasts were tested, chosen for biomechanical features that differ across modes in the training data~\cite{bhakta2020, zhou2025, simon2014}: (i)~LW vs.\ RA at knee stance pre-flexion; (ii)~LW vs.\ RD at knee peak torque. Each contrast was modeled as $\text{outcome} \sim \text{mode} + (\text{mode} \mid \text{subject})$, pooling across ramp grades, with LW data restricted to $0.4$~m/s to match the ramp-trial speed. For within-mode grade scaling, we first fit $\text{outcome} \sim \text{grade} + (\text{grade} \mid \text{subject})$ to the training data. Where the grade effect was significant, the same model was fit to the deployed data, and fixed-effect estimates and $95\%$~CIs were compared side-by-side. Tests used $\alpha = 0.05$ in MATLAB~2025b. With four deployed participants, we report the statistical results as descriptive rather than as adequately powered hypothesis tests, indicating the preliminary capability of the proposed approach.


\section{Results}
\begin{figure}[t]
\centering
\includegraphics[width=\columnwidth]{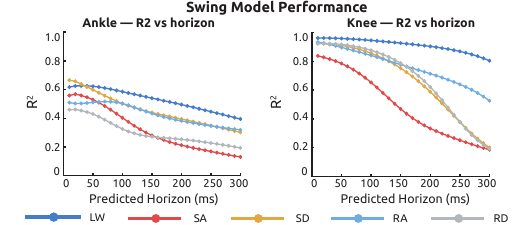}
\caption{Offline swing-phase trajectory estimation R\textsuperscript{2} across prediction horizons for the ankle (left) and knee (right).\label{fig:swing_horizon}}
\end{figure}

\subsection{Torque and Position Estimation Performance}
Offline stance-phase torque was estimated with broadly comparable accuracy across modes ($R^2 = 0.47$--$0.58$ at the knee, $0.37$--$0.59$ at the ankle; Table~\ref{tab:stance_perf}). Swing-phase trajectory $R^2$ decreased with prediction horizon across all modes (Fig.~\ref{fig:swing_horizon}), with the knee maintaining a higher $R^2$ than the ankle across all modes and horizons.

\subsection{Level Walking Adaptation}
\begin{figure*}[t]
\centering
\includegraphics[width=\textwidth]{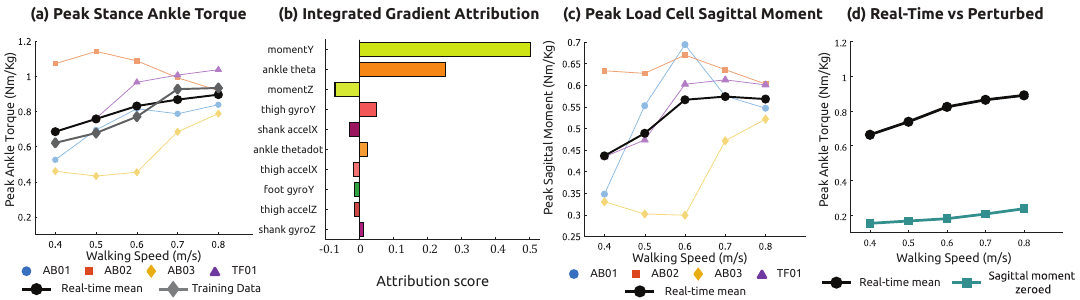}
\caption{(a) Peak stance ankle torque across walking speeds, with subject-level data points and the subject-averaged real-time mean and training-data mean overlaid. (b) Integrated gradient attribution scores for the 10 highest-attributed input channels driving predicted peak ankle torque during LW. (c) Weight normalized peak load cell sagittal moment across walking speeds, with the subject-averaged mean overlaid. (d) Real-time peak ankle torque compared against offline inference with the sagittal load cell moment zeroed (subject-averaged means).\label{fig:adaptation}}
\end{figure*}

The training data exhibited a significant positive scaling of peak ankle
torque with walking speed ($\beta = 0.96$~Nm/kg per m/s, 95\% CI
$[0.42, 1.50]$, $\text{df} = 15.0$, $p = 0.002$). The deployed model
produced a positive slope ($\beta = 0.52$~Nm/kg per m/s, 95\% CI
$[-0.29, 1.33]$, $\text{df} = 3.0$, $p = 0.076$), with an interval spanning
zero driven by high inter-subject variability from AB02
($\beta = -0.47$). Excluding AB02, the only participant new to the device,
gave $\beta = 0.85$~Nm/kg per m/s (95\% CI $[0.63, 1.07]$,
$\text{df} = 3.1$, $p = 0.001$), with its CI fully contained within the
training CI, indicating consistent scaling between the deployed controller
and its demonstrations. Fig.~\ref{fig:adaptation}(a) shows the subject-averaged scaling trend alongside the training-data mean at each speed.

\begin{table}[t]
\caption{Offline Stance Torque Estimation $R^2$ by Mode\label{tab:stance_perf}}
\centering
\resizebox{\columnwidth}{!}{%
\begin{tabular}{lccccc}
\toprule
 & LW & SA & SD & RA & RD \\
\midrule
Knee  & $0.53 \pm 0.21$ & $0.56 \pm 0.18$ & $0.48 \pm 0.23$ & $0.58 \pm 0.22$ & $0.47 \pm 0.19$ \\
Ankle & $0.47 \pm 0.18$ & $0.49 \pm 0.15$ & $0.59 \pm 0.16$ & $0.37 \pm 0.16$ & $0.43 \pm 0.20$ \\
\bottomrule
\end{tabular}%
}
\end{table}

\subsection{Learned Sensor Dependence}
IG was applied to attribute the model's predicted ankle torque to
individual sensor channels. Across all 20 LW trials (4 participants
$\times$ 5 walking speeds), predicted peak ankle torque rose by an average
of 0.714~Nm/kg above its resting-state baseline. IG attributed
$\sim$72\% of this increase to the sagittal-plane load cell moment
($M_y$, mean signed IG $= 0.51$~Nm/kg), followed by the ankle joint angle
(mean signed IG $= 0.26$~Nm/kg); all other channels contributed less than
$0.10$~Nm/kg in absolute magnitude (Fig.~\ref{fig:adaptation}(b)). The
measured peak $M_y$ across participants tracked the peak ankle torque
scaling pattern (Fig.~\ref{fig:adaptation}(c)). Zeroing $M_y$ during
offline inference reduced mean predicted peak ankle torque by 76\%
(Fig.~\ref{fig:adaptation}(d)), consistent with the IG-attributed share.

\subsection{Ramp Adaptation}
\begin{figure}[t]
\centering
\includegraphics[width=\columnwidth]{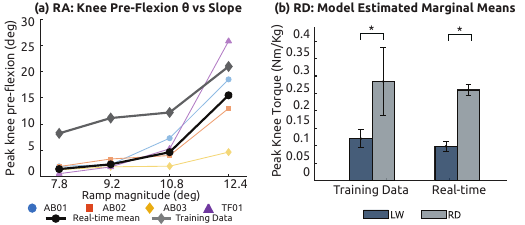}
\caption{Ramp adaptation results. (a) Peak knee pre-flexion vs.\ ramp grade during ramp ascent. Each colored marker denotes one participant's mean peak knee pre-flexion across strides at a given grade. Black denotes the real-time mean across participants; gray denotes the training-data mean across participants at each ramp-grade. (b) Model-estimated marginal means of peak resistive knee torque during LW (blue) and RD (gray) for the training data (left) and real-time deployment (right). Error bars indicate $95\%$ confidence intervals. Asterisks denote statistical significance ($p < 0.05$) between two modes.\label{fig:ramps}}
\end{figure}

Knee pre-flexion was significantly higher in RA than LW ($\beta_{\text{RA-LW}} = 5.03^\circ$, 95\% CI $[1.25, 8.81]$, $df = 5.0$, $p = 0.019$). Within-RA, the training data showed positive grade scaling ($\beta_{\text{slope}} = 3.30^\circ$/deg, 95\% CI $[1.83, 4.77]$, $df = 7.5$, $p < 0.001$), and the deployed model produced grade scaling in the same direction ($\beta_{\text{slope}} = 2.92^\circ$/deg, 95\% CI $[0.53, 5.30]$, $df = 4.0$, $p = 0.027$), corresponding to a $13.4^\circ$ increase across the $7.8$--$12.4^\circ$ tested range. The deployed fixed-effect estimate fell within the training CI. The deployed pre-flexion magnitudes were biased lower than the training reference at each grade (Fig.~\ref{fig:ramps}(a)).

For descent, within-RD grade scaling in the training data was not significant ($\beta_{\text{slope}} = 0.014$~Nm/kg/deg, $df = 9.8$, $p = 0.097$), so the deployed within-RD scaling test was omitted; performance was instead assessed at the mode level. The training data exhibited a significant mode difference in peak knee torque ($\beta_{\text{RD-LW}} = 0.164$~Nm/kg, 95\% CI $[0.051, 0.277]$, $df = 12.7$, $p = 0.008$), which the deployed model reproduced ($\beta_{\text{RD-LW}} = 0.162$~Nm/kg, 95\% CI $[0.137, 0.187]$, $df = 4.3$, $p < 0.001$; Fig.~\ref{fig:ramps}(b)).

\begin{figure*}[t]
\centering
\includegraphics[width=\textwidth]{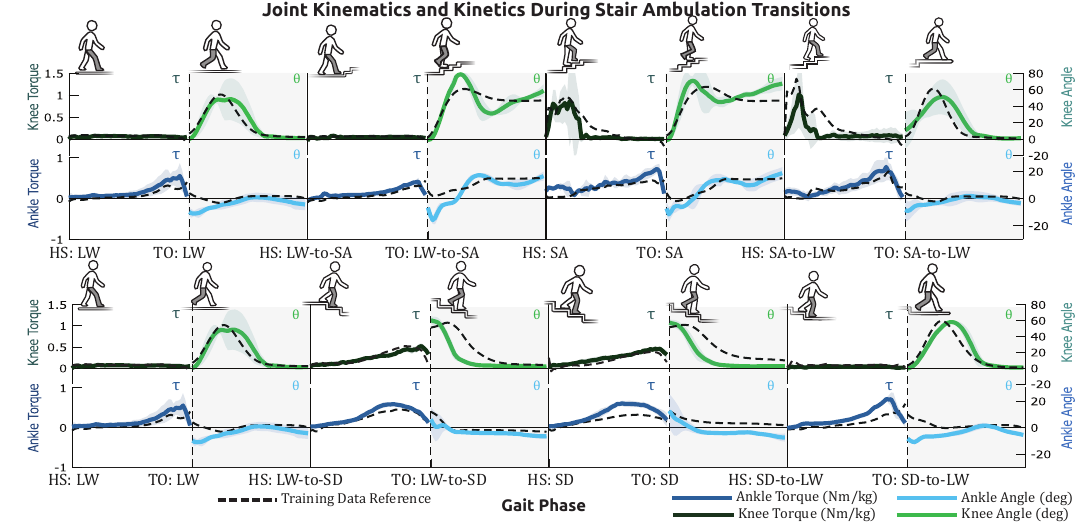}
\caption{Terrain transition behavior for TF01 across full stair circuits. Shaded regions mark swing phases throughout (indicating joint angles, light lines); unshaded regions mark stance phases (estimated joint torque, dark lines). Vertical dashed lines indicate toe-off (TO) events labeled along the x-axis. (Top) Level-walking to stair-ascent and back. (Bottom) Level-walking to stair-descent and back. Knee (green) and ankle (blue) traces are plotted with torque on the left y-axis and angle on the right y-axis per subplot. The reference line is the mean torque and position trajectories from the training dataset for the corresponding stair condition.\label{fig:stair}}
\end{figure*}

\begin{figure*}[t]
\centering
\includegraphics[width=\textwidth]{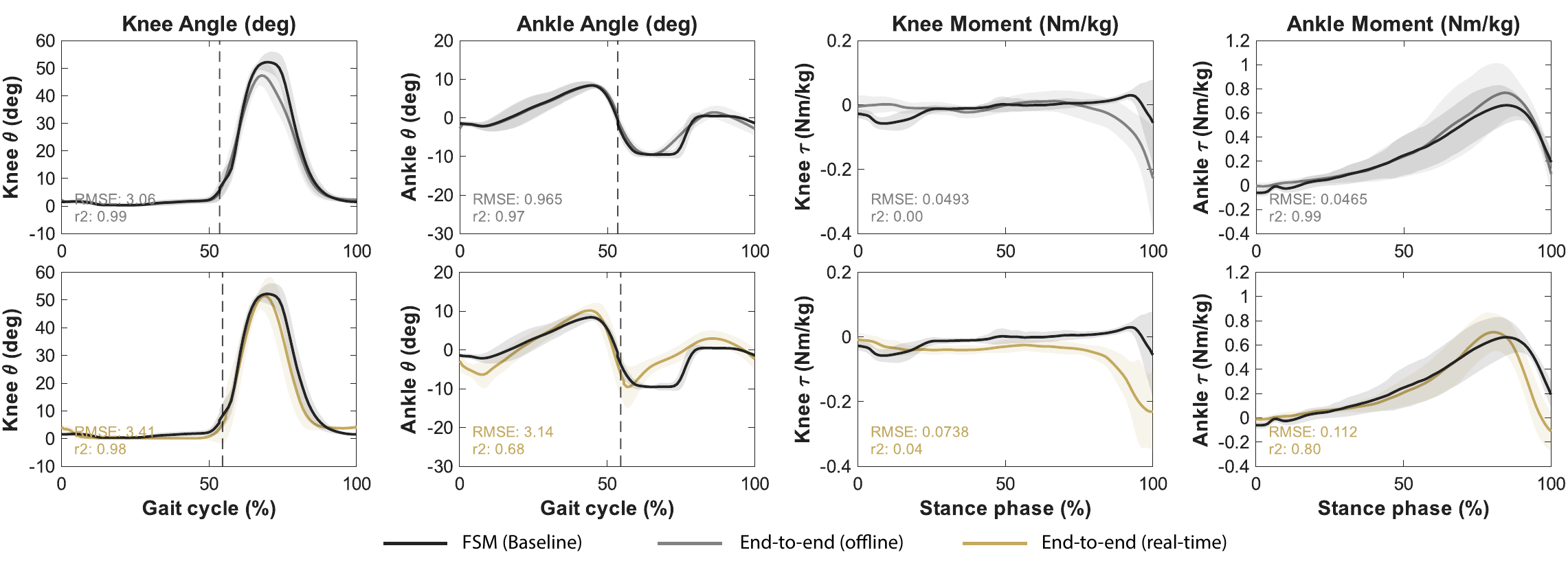}
\caption{Comparison of the proposed end-to-end controller against the FSM baseline: offline prediction (top row) versus real-time deployment (bottom row).\label{fig:compare}}
\end{figure*}

\subsection{Stair Transition}
Fig.~\ref{fig:stair} illustrates transition behavior for TF01 during stair ambulation. As the participant entered SA swing phase, the knee maintained flexion while the ankle dorsiflexed. Upon heel contact, an extension torque was applied at the knee, contributing to push-off. During SD, knee extension torque was generated as the user flexed the knee through stance.

\subsection{Real-Time Deployment Fidelity}
\label{sec:realtime}
Fig.~\ref{fig:compare} shows kinematic and kinetic tracking at the knee and ankle for both conditions relative to the baseline FSM controller.

Knee angle tracking was largely preserved under closed-loop deployment ($R^2 = 0.99$ offline, $0.98$ real-time), with RMSE increasing marginally
(3.06 to 3.14~deg) and peak timing unchanged, indicating that this channel was not substantially affected by closed-loop feedback.

The ankle joint, by contrast, showed strong offline agreement with the FSM reference in both kinematics ($R^2 = 0.97$, RMSE $= 0.965^\circ$) and
kinetics ($R^2 = 0.99$, RMSE $= 0.0465$~Nm/kg), but deviated substantially under real-time deployment (angle: $R^2 = 0.68$, RMSE $= 3.14^\circ$; torque: $R^2 = 0.80$, RMSE $= 0.112$~Nm/kg).

Knee moment agreement with the FSM reference was low in both conditions (offline: $R^2 = 0.00$, RMSE $= 0.0493$~Nm/kg; real-time: $R^2 = 0.04$,
RMSE $= 0.0738$~Nm/kg), driven by a knee flexion torque generated near the late stance-to-swing transition that is absent from the FSM
reference in both conditions. This divergence did not depend on closed-loop deployment, indicating a different source of divergence than real-time distribution shift.
\section{Discussion}
This study evaluated the real-time performance of a powered knee-ankle prosthesis driven by continuous torque and trajectory estimates produced by TCNs. The system was evaluated against two hypotheses: that the controller (i)~reproduces the assistance scaling patterns present in its training data and (ii)~detects environmental context shifts and generates distinguishable joint behaviors across locomotion modes.

A consistent scaling pattern was observed between real-time peak ankle
plantarflexion torque and the corresponding training-data trend across
walking speeds (Fig.~\ref{fig:adaptation}(a)), supporting our first
hypothesis. The deployed slope closely tracked the training scaling,
indicating that the controller inferred a continuous speed context from
raw sensor data and modulated assistance accordingly, with no
per-subject adjustment of the learned policy and no explicit speed estimation. This scaling relationship held for two of the three able-bodied participants (AB01 and AB03) despite the bypass-adapter configuration being absent from training; the exception, AB02, is examined below. TF01's results were consistent with the corresponding pattern in the target population, and these results additionally suggest preliminary generalization of the learned mapping to a small cohort of able-bodied users.

Perturbation analysis revealed that this speed-adapted assistance is mediated by sagittal load cell moment $M_y$. Integrated gradients flagged $M_y$ as a dominant contributor to the model's torque commands (Fig.~\ref{fig:adaptation}(b)) and targeted $M_y$ perturbations confirmed its causal role in ankle torque generation  (Fig.~\ref{fig:adaptation}(d)). Mechanistically, $M_y$ captures axial loading transmitted through the prosthesis and grows with ground reaction force, which scales with walking speed~\cite{nilsson1989}. This makes $M_y$ a biomechanically informative speed cue. AB02's inverted slope provided a direct test of this mechanism: as the only participant naive to the device, AB02 produced atypical $M_y$ loading, and the scaling behavior inverted in proportion. This convergence is notable given that the TCN had no architectural priors encoding biomechanical structure, suggesting deep-learning controllers can develop physically interpretable input dependencies.

The controller produced mode-distinct joint behaviors across ramp conditions (Fig.~\ref{fig:ramps}), supporting our second hypothesis. During ascent, knee pre-flexion was elevated relative to LW and scaled positively with grade, with the deployed slope falling within the training CI --- indicating that the controller recognized grade and modulated its assistance. The deployed pre-flexion magnitudes at heel contact, however, were systematically biased lower than the training reference at each grade, with the offset largest at shallow grades and shrinking as grade steepened (Fig.~\ref{fig:ramps}(a)). We interpret this offset as a consequence of two compounding factors. First, training-data imbalance: LW strides outnumber ramp strides (Fig.~\ref{fig:hardware}(b)), so under MSE optimization the regressor's prediction collapses toward the conditional mean of the training targets, which is dominated by LW; this bias is amplified at shallow grades, where ramp sensor signatures overlap with those of LW. Second, the heterogeneous control policies described in Section~\ref{sec:dataset}, under which some RA conditions did not incorporate stance pre-flexion at heel contact at all, may have pulled the learned target further toward reduced pre-flexion magnitude. A more balanced training distribution, a more homogeneous control policy across constituent protocols, or explicit grade-related input, would likely narrow the magnitude gap. For descent, the deployed knee torque contrast against LW closely matched the training-data contrast (Fig.~\ref{fig:ramps}(b)), demonstrating that the controller replicated the descent assistance shown in the training data.

Stair ambulation provided the strongest demonstration of these capabilities. The transition cue from LW to SA was the user's residual thigh lift, captured by the thigh IMU as they prepared to step onto a stair. This indicates that the controller learned to interpret user intent directly from onboard mechanical sensors, without requiring vision, depth cameras, or other anticipatory environmental sensing. Critically, the transition unfolded continuously rather than via discrete mode switching, demonstrating that complex gait transitions can be produced continuously with a deep learning-based controller.

The real-time comparison in Section~\ref{sec:realtime} surfaced two additional findings meriting discussion. In the offline condition, the TCN model's predictions were computed from FSM-driven sensor logs without closing the control loop, so its own outputs never altered its
future inputs; in the real-time condition, the model's outputs directly drove the prosthesis, and any resulting deviation from the demonstrated trajectory became part of the next timestep's input. This distinction isolates two mechanistically distinct sources of divergence from the FSM reference, and indicates that offline prediction accuracy is a necessary but not sufficient condition for reliable closed-loop deployment.

The ankle showed strong offline agreement with the FSM reference in both kinematics and kinetics, yet deviated substantially once deployed in closed loop. This indicates input distribution shift under real-time deployment: small early deviations from the demonstrated trajectory are fed back as input, compounding over successive timesteps~\cite{bengio2015}. Addressing this drift requires exposing the model to states it would
actually encounter under its own closed-loop behavior. DAgger~\cite{ross2011} and scheduled sampling~\cite{bengio2015} both do this by incorporating the model's own rolled-out predictions into training, narrowing the gap between training and deployment state distributions. 

The knee moment, by contrast, diverged from the FSM reference in both the offline and real-time conditions alike, indicating a cause
independent of closed-loop feedback. The deployed controller generated a flexion torque near the late-stance-to-swing transition that the FSM baseline did not produce. As with the ramp pre-flexion offset described above, this is consistent with heterogeneous, per-subject-tuned training targets pulling the learned behavior toward a conditional mean; because the FSM baseline used here reflects one of the many demonstrated controllers, a clean one-to-one comparison of target behavior was not possible for this channel. This
underscores that a fixed, consistent target behavior is necessary not only for the network to learn a well-defined control policy, but for any single baseline to serve as a meaningful reference for evaluating it.

Several limitations should be acknowledged. First, the controller's behavior reflects the assistance patterns embedded in its training data, which were generated by controllers individually tuned per subject rather than a single consistent control law (Section~\ref{sec:dataset}). Its performance is therefore bounded by demonstration consistency: disagreement between these per-subject targets propagates into the deployed controller, as observed in both the RA pre-flexion offset and the late-stance knee moment divergence described above. Future work would benefit from training on more uniform demonstrations that more closely approximate able-bodied biomechanics, providing a higher ceiling for the controller to learn from.

A second limitation is that the evaluation cohort was both small and compositionally imbalanced: three of the four real-time participants were able-bodied individuals ambulating via a bypass adapter, and only one had transfemoral amputation. This configuration differs from natural prosthesis use in load transfer and residual-limb control, and AB02's atypical $M_y$ loading illustrates that this difference can be substantial for a participant naive to the device. Therefore, these findings should be viewed as preliminary evidence of potential for end-to-end prosthetic control, pending replication in a larger and more representative cohort.
\section{Conclusion}
This study demonstrated real-time closed-loop control of a powered
knee-ankle prosthesis across five locomotion modes, using a deep learning control architecture with no impedance parameterization, intent classifier, or
per-subject adjustment of the learned policy. The controller adapted its assistance to walking speed and
locomotion mode. Integrated gradients and targeted perturbation showed that the model converged on a
biomechanically interpretable input dependence without architectural priors, indicating that end-to-end deep controllers can develop
physically grounded assistance strategies. These preliminary results
represent a step towards unified, mode-adaptive prosthetic assistance
with substantially reduced tuning burden.

\section*{Acknowledgment}
The authors would like to thank Chase Sun for mechatronics support, and Caesar Guo and the Prosthesis Vertically Integrated Project (VIP) members for assistance with experiments.

\section*{Funding Data}
\begin{itemize}
\item DoD CDMRP (Award No.\ W81XWH2211091)
\item NIH Health Director New Innovator (Award No.\ DP2-HD111709)
\end{itemize}

\section*{Code Availability}
The TCN architecture and the trained stance (torque) and swing (trajectory) models used in the real-time experiments, including the input normalization statistics for each model, are publicly available under the MIT License at \url{https://github.com/yohanshim1/end-to-end-prosthesis-controller}.

\bibliographystyle{asmejour}
\bibliography{references}

\end{document}